\documentclass{EG-style/egpubl}
\usepackage{EG-style/eg2026}
 
\ShortPresentation      
\usepackage[T1]{fontenc}
\usepackage{EG-style/dfadobe}  

\usepackage{cite}  
\BibtexOrBiblatex
\electronicVersion
\PrintedOrElectronic
\ifpdf \usepackage[pdftex]{graphicx} \pdfcompresslevel=9
\else \usepackage[dvips]{graphicx} \fi

\usepackage{EG-style/egweblnk} 
\graphicspath{{EG-style/}}

\usepackage{multirow}
\usepackage{amsmath}
\usepackage{cuted}
\usepackage{array}
\usepackage{booktabs}
\usepackage[table]{xcolor}
\usepackage{tikz}
\usetikzlibrary{positioning}
\usepackage{cleveref}

\usepackage{arydshln}

\title[2D-SuGaR]%
      {2D-SuGaR: Surface-Aware Gaussian Splatting for Geometrically Accurate Mesh Reconstruction}

\author[Prajwal Gupta C.\,R., Divyam Sheth, Jinjoo Ha, Mirela Ostrek \& Justus Thies]
{\parbox{\textwidth}{\centering 
Prajwal Gupta C.\,R.$^{\dagger\,1}$\orcid{0000-0002-4846-1733}, 
Divyam Sheth$^{\dagger\,1,2}$\orcid{0009-0007-0859-6901}, 
Jinjoo Ha$^{1}$\orcid{0009-0000-0224-1850}, 
Mirela Ostrek$^{1,3}$\orcid{0009-0009-9987-646X} and 
Justus Thies$^{1,2,3}$\orcid{0000-0002-0056-9825}
}
        \\
{\parbox{\textwidth}{\centering $^1$TU Darmstadt, $^{2}$ ELIZA,
         $^3$ Max Planck Institute for Intelligent Systems, $^{\dagger}$ equal contribution 
       }
       }
}

\begin{document}


\teaser{
    \vspace{-0.65cm}
    \includegraphics[width=0.87\linewidth]{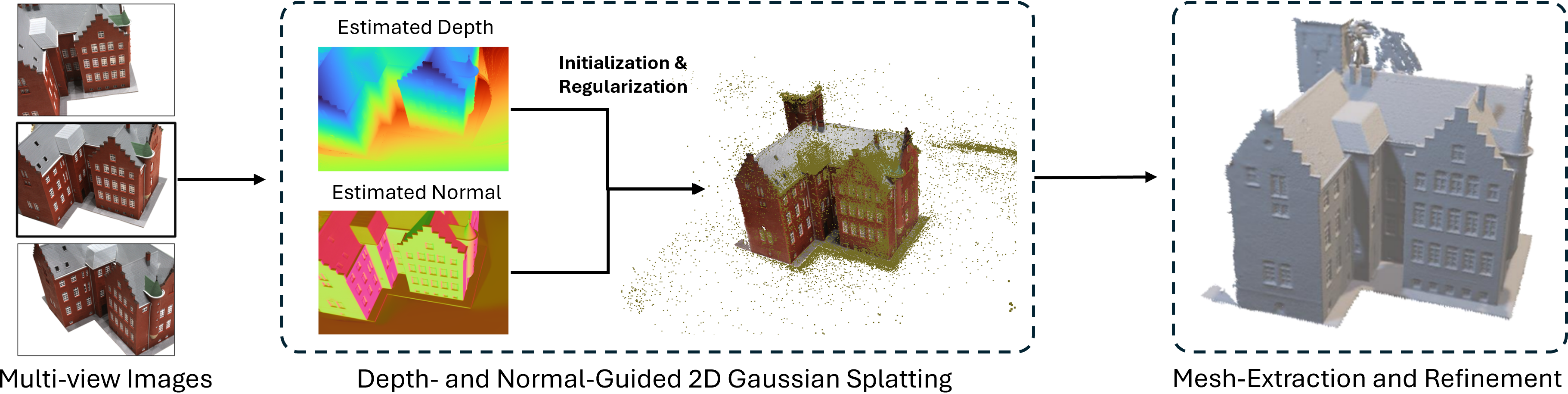}
     \centering
      \caption{
        \textbf{Overview.}
        We propose 2D-SuGaR, a method to reconstruct 3D meshes from multi-view input images by leveraging 2D Gaussian Splatting. We initialize and regularize the 2D Gaussian primitives with pretrained normal and depth priors, and extract a mesh from this volumetric representation. We refine the mesh afterwards with additional re-rendering losses.
      }
    \label{fig:teaser}
}

\maketitle
\begin{abstract}
3D Gaussian Splatting enables the reconstruction of a volumetric scene representation from multi-view images that allows for real-time novel-view point synthesis, however, it struggles with recovering an accurate surface geometry.
While 2D Gaussian Splatting (2DGS) addresses this through surface-aligned primitives, its performance depends critically on the initialization quality.
Reliance on Structure-from-Motion (SfM) limits the initialization flexibility as well.
In this work, we present two key contributions to enhance 2DGS and the extraction of a clean surface mesh.
First, we incorporate monocular depth and normal priors for robust initialization, coupled with a clustering-based pruning strategy to eliminate degenerate Gaussians.
Second, we introduce a joint mesh-Gaussian refinement similar to SuGaR, that relaxes the strict 2D constraint by transitioning to 3D primitives, providing stronger training signals.
Evaluated on the DTU dataset, our method achieves state-of-the-art mesh
reconstruction with a Chamfer Distance of 0.67, outperforming prior methods.
%


\end{abstract}  


%
\section{Introduction}
Reconstructing accurate 3D meshes from multi-view images remains a fundamental challenge in computer vision, with applications in AR/VR, robotics, and digital content creation.
Recent advances based on Gaussian Splatting~\cite{kerbl2023,huang2024} have enabled fast optimization and real-time photorealistic rendering.
%
These methods represent a 3D scene in a volumetric way by a set of translucent Gaussians.
The Gaussians are optimized using a differentiable renderer, where rendering proceeds by splatting and rasterizing the Gaussians in screen space.
It is worth noting that this scheme is sensitive to the initialization of Gaussians.
Extracting a mesh-based representation from optimized Gaussians for usage in classical computer graphics frameworks is typically performed as a post-processing step, preventing errors introduced at this stage from being corrected.
This limits the ability to capture accurate geometry and fine details.
Our work builds on top of 2D Gaussian-Splatting~\cite{huang2024} to reconstruct meshes from multi-view input images.
To address the challenges mentioned above, we propose several contributions.
First, we incorporate monocular depth and normal priors to enable \textit{robust initialization and regularization during optimization}, which proves useful given the highly unconstrained nature of 3D reconstruction.
We also introduce a \textit{clustering-based pruning method to remove spurious Gaussians} that are not eliminated during adaptive density control. 
Finally, we propose a \textit{joint mesh refinement stage} taking inspiration from SuGaR~\cite{guedon2023sugar}. 
%

\section{Method}
%
%
We leverage 2D Gaussian primitives as an intermediate representation~\cite{huang2024} to recover a 3D mesh from multi-view input images, see \Cref{fig:teaser}.
To initialize the 2D Gaussians, we augment feature points from Structure-from-Motion (SfM) with estimates from monocular depth and normal predictions~\cite{hu2024metric3d}.
Besides initialization, the normal predictions from this general prior, are used in a new normal loss that encourages surface alignment of the 2D Gaussians.
Once the 2D Gaussian-based representation is optimized, we extract a coarse mesh and further refine it using joint mesh-gaussian optimization. ~\cite{guedon2023sugar}. 
%

%
%


\subsection{Depth- and Normal-Guided 2DGS}
\label{subsec:init}
\begin{figure}[t]
    \centering
        \centering
        \includegraphics[width=\linewidth]{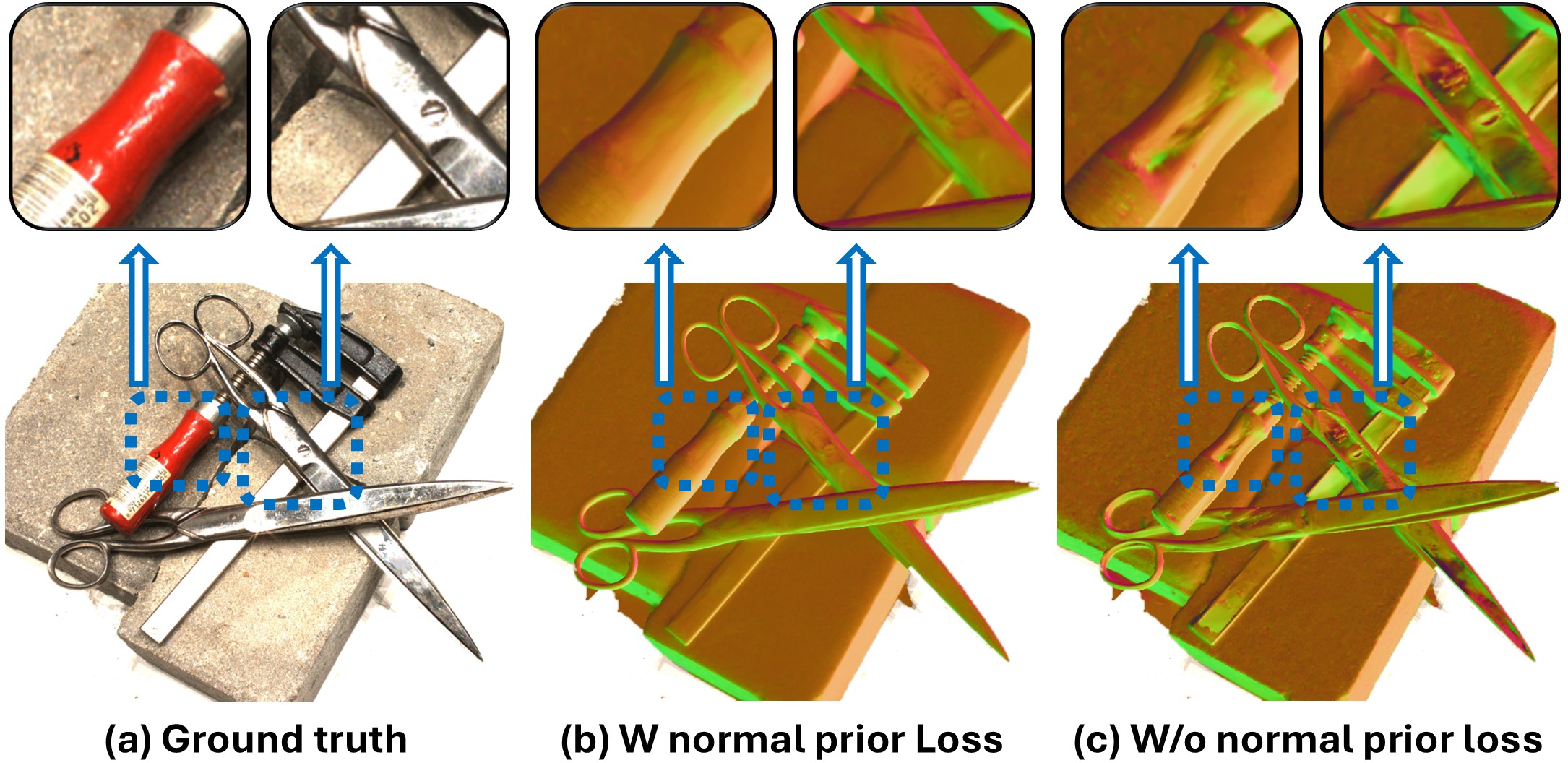}
        \label{fig:normal_gt}
    \caption{
    \textbf{Effect of normal loss.}
    Our new normal loss reduces 2DGS reconstruction artifacts, especially in specular regions, and leads to a smoother and more faithful representation of the surface.
    }
    \label{fig:normal_comparison}
\end{figure}

2DGS~\cite{huang2024} is dependent on the initialization through SfM.
However, SfM fails in textureless and occluded regions, leaving gaps where reconstruction is most challenging.
We address this by supplementing the SfM points with monocular depth and normal estimates from Metric3D~\cite{hu2024metric3d}.
To resolve the scale ambiguity of the depth, we compute per-image scale correction using SfM as reference.
We then sample 1000 pixels per image over a 2D Gaussian distribution (50\% in center) and backproject the depth to 3D space to initialize Gaussians, with orientations from the normal estimates and colors from the input image.
Based on this initialization, we optimize the 2D Gaussian primitives using the loss:
\begin{equation}
    \mathcal{L} = \mathcal{L}_p + \lambda_c \mathcal{L}_c + \lambda_n \mathcal{L}_n + \lambda_d \mathcal{L}_d .
\end{equation}
Here, $\mathcal{L}_p$ denotes the photometric loss combining an $\ell_{1}$ term with the D-SSIM term from~\cite{kerbl2023}. $\mathcal{L}_c$ is the normal consistency loss, and $\mathcal{L}_d$ is the depth distortion loss (refer to Sec. 1 of the supplementary material).
While the original depth-normal consistency loss $\mathcal{L}_c$ ensures alignment between the rendered depth and rendered normals, we introduce a new term leveraging the monocular normal estimates $\mathbf{\Tilde{N}}$:
\begin{equation}
    \mathcal{L}_n = 1 - \mathbf{N}\cdot\mathbf{\Tilde{N}} .
\end{equation}
The normal prior loss leads to normal maps that better reflect the underlying surface, as illustrated in \Cref{fig:normal_comparison}. 
Note that errors in depth estimates may result in the initialization of spurious Gaussians. These are further amplified by the adaptive density control strategy, creating isolated Gaussian islands that produce undesirable reconstruction artifacts.
To address this, we introduce a \emph{clustering-based pruning strategy} using DBSCAN~\cite{DBSCAN}. The neighborhood radius (\(\epsilon\)) is dynamically determined using the $k$-distance heuristic. For each Gaussian $p_i$, we compute the distance $d_i^{(k)}$ to its $k$-th nearest neighbor. We set $\epsilon$ to the 90\textsuperscript{th} percentile of $d_i^{(k)}$.
%
We retain only the largest cluster, assuming that true geometry forms a single connected component.
%


\subsection{Mesh Extraction and Refinement}
\label{subsec:refinement}
Post optimization, we render depth maps from the training views. We fuse them into a Truncated Signed Distance Field (TSDF) and apply marching cubes to extract an initial mesh. We then instantiate new 3D Gaussians on the extracted mesh, following SuGaR~\cite{guedon2023sugar}. We bind a single thin 3D Gaussian to each triangle of the mesh, sampled on the surface of the triangle. To ensure that the Gaussians remain bound to their respective triangles during optimization, we specify the Gaussian means in barycentric coordinates with respect to the mesh vertices. We refine it using:
\begin{equation}
\mathcal{L}_{r} = \mathcal{L}_{p} + \gamma \mathcal{L}_{Lap} + \delta \mathcal{L}_{m} ,
\end{equation}
%
where $\mathcal{L}_{Lap}$ and $\mathcal{L}_{m}$ are the standard mesh laplacian smoothing and mesh normal consistency losses respectively. 
Note that since the Gaussians in 2DGS have only two scale parameters and SuGaR is based on 3DGS, we introduce an additional dimension with a small initial value of $0.001$.
This modification aligns well with the final goal of joint refinement, where one dimension of each Gaussian is progressively reduced to produce a flat, surface-aligned shape.
%

%

\subsection{Implementation Details} 
The proposed method builds on \cite{huang2024} and \cite{guedon2023sugar}, implemented in PyTorch with custom CUDA kernels for efficient execution.
All our experiments are run on a single RTX 3090 GPU. 

\smallskip
\noindent
\textbf{Depth- and Normal-Guided 2DGS.}
We adopt the training configuration used in 2DGS to ensure a fair and direct performance comparison.
Each scene is optimized over 30K iterations, where we set $\lambda_c = 0.05$, $\lambda_n = 0.1$ and $\lambda_d = 1000$.
During the depth-prior based initialization, we sample 1000 pixels per image in the scene.
For each sampled pixel, we compute its corresponding 3D position using the monocular depth estimate and initialize a 2D Gaussian at that point. 
For scenes with very large number of images, the total number of samples is capped at 75000, with equal number of samples from each image.
\begin{equation}
N_{\text{samples}} = \min(75000,\, 1000 \times N_{\text{images}})
\end{equation}
When using the depth-prior based initialization, we apply our clustering-based pruning after 7K iterations using DBSCAN with $\text{min\_samples} = k = 4$ and $\epsilon = P_{90}(\{d_i^{(k)}\}_{i=1}^N)$. 
%



\smallskip
\noindent
\textbf{Mesh Refinement.}
%
%
We refine the extracted mesh for 8K iterations which takes approximately 8 minutes on a single RTX 3090.
Even though we get good results in just 4K iterations, we got the best overall results for 8K iterations. 
15K iterations did not improve the performance significantly for the time overhead it added.
We set $\gamma = 5.0$ and $\delta = 0.1$ for all scenes as default values used in ~\cite{guedon2023sugar}.
%
%
%

\begin{figure*}[!ht]
    \centering
    \includegraphics[width=\textwidth ]{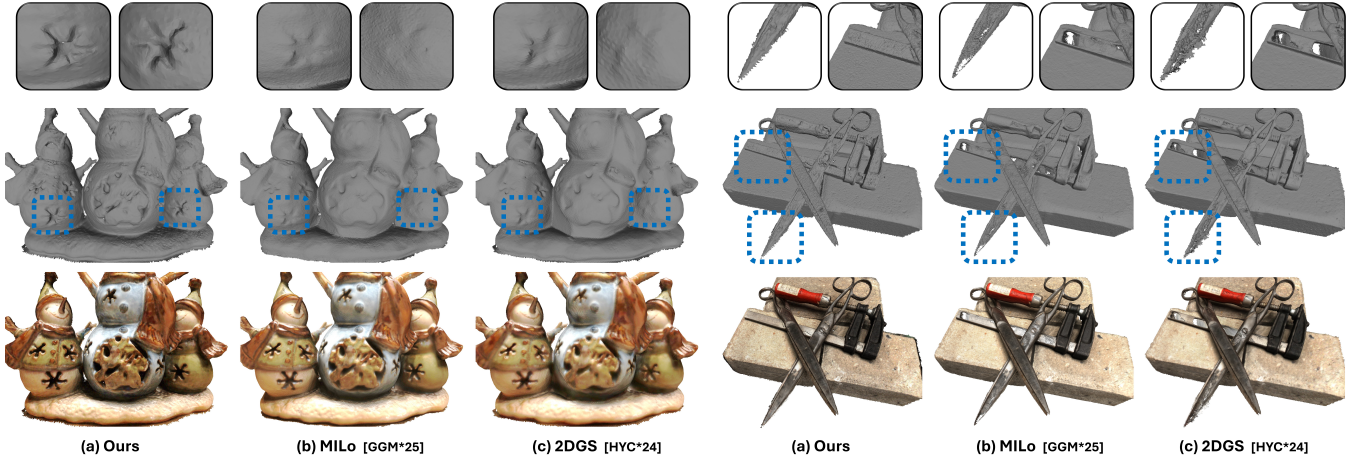}
    \caption{\textbf{Qualitative comparison of meshes on the DTU benchmark.} 
    Our method produces more detailed and complete meshes as can be seen in the zoom-ins.   
    }
    \label{fig:main_results}
\end{figure*}

\definecolor{highcolor}{RGB}{255, 178, 178}    
\definecolor{midcolor}{RGB}{255, 217, 178}   
\definecolor{lowcolor}{RGB}{255, 255, 178}   
\newcolumntype{P}[1]{>{\raggedright\arraybackslash}m{#1}} 
\newcolumntype{C}[1]{>{\centering\arraybackslash}m{#1}}  

\begin{table*}[!htb]
    \centering
    \footnotesize
    
    \begin{tabular}{P{2.5cm}C{0.4cm}C{0.4cm}C{0.4cm}C{0.4cm}C{0.4cm}C{0.4cm}C{0.4cm}C{0.4cm}C{0.4cm}C{0.4cm}C{0.4cm}C{0.4cm}C{0.4cm}C{0.4cm}C{0.4cm}C{0.7cm}C{0.7cm}}
    \toprule
    \textbf{Method} & 24 & 37 & 40 & 55 & 63 & 65 & 69 & 83 & 97 & 105 & 106 & 110 & 114 & 118 & 122 & \textbf{Mean}$ {\scriptscriptstyle \downarrow} $ & \textbf{Time}$ {\scriptscriptstyle \downarrow} $ \\
    \cmidrule(r){1-1} \cmidrule(r){2-16} \cmidrule(l){17-18}
    NeRF \cite{mildenhall-2020} & 1.90 & 1.60 & 1.85 & 0.58 & 2.28 & 1.27 & 1.47 & 1.67 & 2.05 & 1.07 & 0.88 & 2.53 & 1.06 & 1.15 & 0.96 & 1.49 & >12h \\
    VoISDF \cite{yariv-2021} & 1.14 & 1.26 & 0.81 & 0.49 & 1.25 & \cellcolor{lowcolor} 0.70 & 0.72 & 1.29 & 1.18 & 0.70 & 0.66 & 1.08 & 0.42 & 0.61 & 0.55 & 0.86 & >12h \\
    NeuS \cite{wang-2021} & 0.83 & 0.98 & 0.56 & \cellcolor{highcolor} 0.37 & 1.13 & \cellcolor{highcolor} 0.59 & \cellcolor{highcolor} 0.60 & 1.45 & \cellcolor{highcolor} 0.95 & 0.78 & \cellcolor{lowcolor} 0.52 & 1.43 & \cellcolor{highcolor} 0.36 & \cellcolor{highcolor} 0.45 & \cellcolor{lowcolor} 0.45 & 0.77 & >12h \\
    3DGS \cite{kerbl2023} & 2.14 & 1.53 & 2.08 & 1.68 & 3.49 & 2.21 & 1.43 & 2.07 & 2.22 & 1.75 & 1.79 & 2.55 & 1.53 & 1.52 & 1.50 & 1.96 & \cellcolor{lowcolor}11.2m \\
    SuGaR \cite{guedon2023sugar} & 1.47 & 1.33 & 1.13 & 0.61 & 2.25 & 1.71 & 1.15 & 1.63 & 1.62 & 1.07 & 0.79 & 2.45 & 0.98 & 0.88 & 0.79 & 1.33 & 65m \\
    2DGS \cite{huang2024} & \cellcolor{lowcolor} 0.48 & 0.91 & 0.39 & \cellcolor{lowcolor} 0.39 & \cellcolor{lowcolor} 1.01 & 0.83 & 0.81 & 1.36 & 1.27 & 0.76 & 0.70 & 1.40 & \cellcolor{lowcolor} 0.40 & 0.76 & 0.52 & 0.80 & \cellcolor{midcolor}10.9m \\
    GOF \cite{yu-2024} & 0.50 & 0.82 & 0.37 & \cellcolor{highcolor} 0.37 & 1.12 & 0.74 & 0.73 & \cellcolor{midcolor} 1.18 & 1.29 & 0.68 & 0.77 & \cellcolor{lowcolor} 0.90 & 0.42 & 0.66 & 0.49 & 0.74 & 75m \\
    RaDe-GS \cite{zhang-2024} & \cellcolor{midcolor} 0.46 & \cellcolor{lowcolor} 0.73 & \cellcolor{midcolor} 0.33 & \cellcolor{midcolor} 0.38 & \cellcolor{highcolor} 0.79 & 0.75 & 0.76 & \cellcolor{lowcolor} 1.19 & 1.22 & \cellcolor{midcolor} 0.62 & 0.70 & \cellcolor{highcolor} 0.78 & 0.68 & \cellcolor{midcolor} 0.47 & 0.68 & \cellcolor{midcolor} 0.68 & 11.5m \\
    MILo \cite{guedon-2025} & \cellcolor{highcolor} 0.43 & 0.74 & \cellcolor{lowcolor} 0.34 & \cellcolor{highcolor} 0.37 & \cellcolor{midcolor} 0.80 & 0.74 & 0.70 & 1.21 & 1.22 & 0.66 & 0.62 & \cellcolor{midcolor} 0.80 & \cellcolor{midcolor} 0.37 & 0.76 & 0.48 & \cellcolor{midcolor} 0.68 & 35m \\
    Ours w/o Refinement & 0.50 & \cellcolor{midcolor} 0.67 & \cellcolor{midcolor} 0.33 & 0.44 & 1.24 & \cellcolor{lowcolor} 0.70 & \cellcolor{lowcolor} 0.63 & 1.24 & \cellcolor{lowcolor} 1.13 & \cellcolor{lowcolor} 0.63 & \cellcolor{midcolor} 0.49 & 1.29 & 0.43 & 0.56 & \cellcolor{midcolor} 0.44 & \cellcolor{lowcolor} 0.72 & \cellcolor{highcolor}10.5m \\
    Ours & \cellcolor{lowcolor} 0.48 & \cellcolor{highcolor} 0.65 & \cellcolor{highcolor} 0.29 & \cellcolor{lowcolor} 0.39 & 1.24 & \cellcolor{midcolor} 0.69 & \cellcolor{midcolor} 0.61 & \cellcolor{highcolor} 1.17 & \cellcolor{midcolor} 1.10 & \cellcolor{highcolor} 0.56 & \cellcolor{highcolor} 0.45 & 1.19 & \cellcolor{midcolor} 0.37 & \cellcolor{lowcolor} 0.50 & \cellcolor{highcolor} 0.40 & \cellcolor{highcolor} 0.67 & 18.5m \\
    \bottomrule
    \end{tabular}
    \caption{\label{tab:DTU_benchmark}
    \textbf{Quantitative comparison of the mesh reconstruction quality of state-of-the-art methods on 15 scenes from the DTU Dataset.} We report the average Chamfer Distance between the ground truth mesh and the reconstructed mesh, as well as the reconstruction time.}
\end{table*}

\newpage
\section{Results}


We evaluate on 15 DTU scenes~\cite{jensen2014large}, using masks for bounded reconstructions following~\cite{huang2024}, with 49 or 64 images per scene.
In \Cref{tab:DTU_benchmark}, we compare against several state-of-the-art approaches based on the Chamfer Distance (CD) and training time.
Some of these works learn an implicit representation of the scene \cite{mildenhall-2020},  \cite{yariv-2021}, \cite{wang-2021}, while others learn an explicit representation of the scene  \cite{kerbl2023}, \cite{huang2024}, \cite{guedon2023sugar}.
Our method achieves the lowest Chamfer Distance among all compared approaches.
However, the mesh refinement stage is time-consuming compared to other explicit representation methods.
Even without the mesh refinement stage, our method remains more accurate and is also faster than other methods.
The final scene representation also contains fewer Gaussians than ~\cite{huang2024}.
In \Cref{fig:main_results}, we show a qualitative comparison to 2DGS and the concurrent work MILo~\cite{guedon-2025} which also aims at recovering meshes from multi-view images.
As shown, our method recovers more detailed and smooth surface representations without noticeable boundary artifacts. Other methods oversmooth the surface and compensate by baking details into the texture. In contrast, our approach preserves geometric detail in the mesh itself (see zoom-ins in the top row). Furthermore, our meshes are more complete, with fewer holes and gaps in the reconstructed geometry (e.g. the scissors).
%


\smallskip
\noindent
\textbf{Ablations -- Depth- and Normal-Guided 2DGS.} 
In \Cref{tab:2DGS_ablation}, we study the impact of each new component by sequentially adding them to 2DGS and evaluating the quality of mesh reconstruction using the DTU dataset.
2DGS + Normal Init. refers to the 2DGS pipeline where the orientation of the Gaussians is initialized using the normal priors, resulting in a modest performance improvement.
Next, we incorporate the normal prior loss leading to a reasonable improvement in performance.
The depth-prior based initialization only results in a modest improvement in performance on the DTU dataset.
However, it can be highly effective in scenes where it is difficult to get a good initialization through SfM.
Finally, we add the mesh refinement step which produces a significant performance improvement.

\begin{figure}[t!]
    \centering
    \includegraphics[width=\columnwidth ]{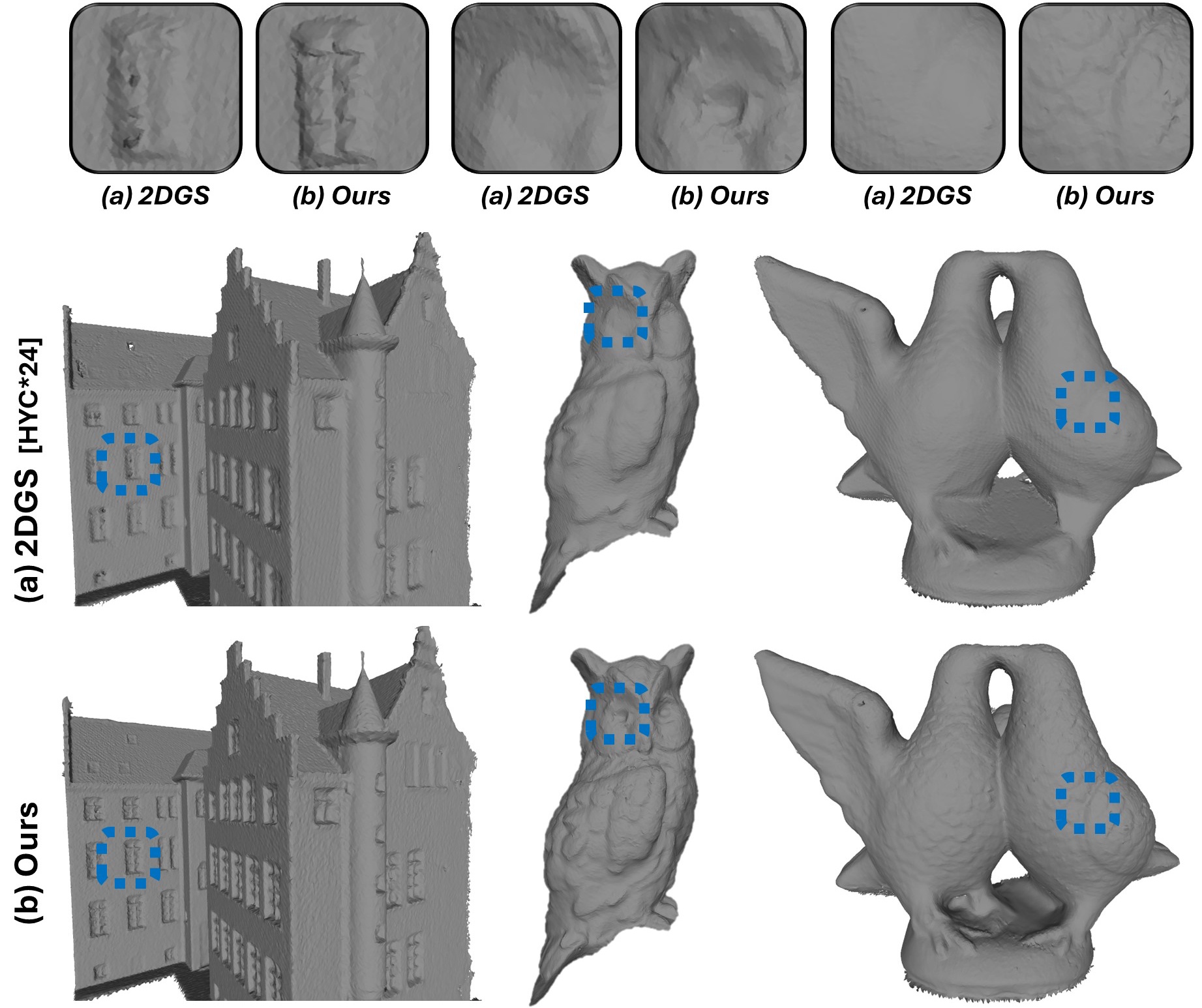}
    \vspace{-0.15cm}   
    \caption{
    \textbf{Qualitative comparison to 2DGS.} Our meshes have fine geometric details such as windows, eyes, and feathers.}
    \vspace{-0.1cm}   
    \label{fig_comparison}
\end{figure}

\vspace{-0.5em}
\noindent
\textbf{Ablations -- Mesh Refinement.}
In \Cref{tab:refinement}, we compare two joint refinement strategies applied on top of vanilla 2DGS meshes to validate our relaxation approach.
Relaxed 3D adds a small third dimension ($s_z = 10^{-3}$) to 2D surfels and refines using 3D Gaussian rasterization with losses from~\cite{kerbl2023} and~\cite{guedon2023sugar}.
Strict 2D maintains rigid surfels using a surfel rasterizer.
Without additional losses, results degrade as primitives drift along depth axis.
With the regularization losses from  ~\cite{huang2024}, it performs similar to the baseline but underperforms compared to Relaxed 3D.
The results confirm our key insight that pure 2D primitives over-constrain mesh adaptation, while the small third dimension enables surface-normal adjustments without losing alignment.

\begin{table}[h!]

    \small
    \centering
    \footnotesize
    \begin{tabular}{lccc}
    \toprule
    \textbf{Added Component} & \textbf{Accuracy} $ {\scriptscriptstyle \downarrow} $ & \textbf{Completion} $ {\scriptscriptstyle \downarrow} $ & \textbf{Average} $ {\scriptscriptstyle \downarrow} $ \\
    \midrule
    2DGS (Base) & 0.701 & 0.818 & 0.759 \\
    + Normal Init. & 0.707 & 0.803 & 0.755 \\
    + Normal Prior Loss & 0.692 & 0.749 & 0.72 \\
    + Depth Init. & 0.689 & 0.741 & 0.715 \\
    + Mesh Refinement & 0.655 & 0.693 & 0.674 \\
    \bottomrule
    \end{tabular}
    \vspace{-0.15cm}   
    \caption{
        \textbf{Quantitative ablation study evaluating the contribution of each component of our method for mesh extraction.}
    }
    \vspace{-0.1cm}   
    \label{tab:2DGS_ablation}
\end{table}

\begin{table}[h!]
    \small
    \centering
    \begin{tabular}{@{}lcc@{}}
    \toprule
    \textbf{Method}  & \textbf{Mean CD} $ {\scriptscriptstyle \downarrow} $ \\
    \midrule
    2DGS (Baseline) & 0.759 \\
    \hdashline
    Relaxed 3D (Ours)& \textbf{0.713} \\
    Strict 2D (Base)  & 0.787 \\
    Strict 2D (Regularized) & 0.756 \\
    \bottomrule
    \end{tabular}
    \vspace{-0.15cm}   
    \caption{
    \textbf{Mesh refinement with different primitives.} }
    \vspace{-0.1cm}  
    
    \label{tab:refinement}
\end{table}
\section{Conclusion}

We have proposed a novel mesh reconstruction scheme based on 2D and 3D Gaussian Splatting.
We improve the 2DGS~\cite{huang2024} reconstruction quality using off-the-shelf monocular depth and normal estimation models for initialization of the Gaussians as well as regularization during optimization.
Based on the optimized 2DGS representation, we extract a mesh and jointly refine it by constraining Gaussians to the mesh surfaces~\cite{guedon2023sugar}.
Through a series of experiments, we show the strengths and weaknesses of each introduced methodological aspect of our proposed method and are able to demonstrate superior quality in comparison to state-of-the-art approaches.
%
%

In the future, it will be interesting to investigate the effects of additional depth-based regularization. Improving the adaptive density control strategy is another exciting direction that may lead to better reconstructions.

\textbf{Acknowledgment } Divyam Sheth is supported by \href{https://eliza.tu-darmstadt.de}{ELIZA} (Konrad Zuse School of Excellence) via \href{https://www.daad.de}{DAAD}, funded by the Federal Ministry of Education and Research. The project is supported by the ERC Starting Grant 101162081 ``LeMo'' and the DFG Excellence Strategy— EXC-3057, Project No. 533677015.

 
\bibliographystyle{EG-style/eg-alpha-doi} 
\bibliography{main}       



\end{document}


\maketitle

\section{Loss Functions}

During optimization of the 2D Gaussians, we combine a photometric loss with a variety of geometric regularization losses for accurate reconstruction and faithful surface-alignment of the Gaussians.

\textit{Photometric Loss.}\quad We use the same photometric loss used in 3DGS~\cite{kerbl2023}, which is a combination of an $L1$ term and a D-SSIM term.
\begin{equation}
    \mathcal{L}_p = 0.8 \cdot \mathcal{L}_1 + 0.2 \cdot \mathcal{L}_{D-SSIM}
\end{equation}

\textit{Depth-Normal Consistency Loss.}\quad This is same as the normal consistency loss used in 2DGS~\cite{huang2024} to ensure consistency between the rendered depth and the rendered normals. Since each ray passes through multiple semi-transparent 2D Gaussians, the depth rendered at each pixel is considered to be at the median point of intersection where the accumulated opacity reaches $0.5$. The loss is formulated as:
\begin{equation}
    \mathcal{L}_c = \sum_i \omega_i(1-\mathbf{n}_i\cdot\mathbf{N})
\end{equation}
where $i$ indexes over the Gaussians intersecting with the ray, $\omega_i = \alpha_i \mathcal{G}_i(\mathbf{u}(\mathbf{x})) \Pi_{j=1}^{i-1}(1-\alpha_j \mathcal{G}_j(\mathbf{u}(\mathbf{x})))$ is the volumetric rendering weight, $\mathbf{n}_i$ is the normal of the $i^{th}$ splat and $\mathbf{N}$ is the normal estimated from the gradients of the rendered depth map.

\textit{Normal Prior Loss.}\quad This term leverages the monocular normal estimates $ \tilde{N} $ to ensure the rendered normals accurately represent the surface orientation. While the depth–normal consistency loss enforces agreement between rendered depth and rendered normals, it does not guarantee alignment of Gaussians with the true surface. This can occur when the rendered depth and rendered normals are consistently incorrect. To avoid such degenerate cases, we combine the depth–normal consistency loss with a normal prior loss. When the rendered normals are accurate, reducing the depth-normal consistency loss drives the Gaussians towards the true surface. The loss is formulated as:
\begin{equation}
    \mathcal{L}_n = 1 - \mathbf{N}\cdot\mathbf{\tilde{N}}
\end{equation}

\textit{Depth Distortion Loss.}\quad Similar to 2DGS~\cite{huang2024}, we use the depth distortion loss to prevent the spreading of Gaussians, ensuring a tight alignment with the surface.
\begin{equation}
    \mathcal{L}_d = \sum_{i,j} \omega_i \omega_j |z_i - z_j|
\end{equation}
where $z_i$ is the intersection depth with the $i^{th}$ Gaussian along the ray.

The final loss function is defined as:
\begin{equation}
    \mathcal{L} = \mathcal{L}_p + \lambda_c \mathcal{L}_c + \lambda_n \mathcal{L}_n + \lambda_d \mathcal{L}_d
\end{equation}
In our experiments, we set $\lambda_c = 0.05$, $\lambda_n = 0.1$ and $\lambda_d = 1000$

\section{Adaptive Density Control and Cluster-Based Pruning}
During the optimization, we observed that the adaptive density control strategy, when combined with depth-prior initialization, can produce undesirable artifacts. One of the underlying causes is the initialization of Gaussians away from the true surface. This could be caused by incorrect monocular depth estimates, inaccurate scale correction or inconsistent scale factor across the image. While we observed that the per-point scale factor usually has a low variance within an image, there can be cases where it differs between objects, such as foreground and background objects having a distinct scale factor. This results in the initialization of spurious Gaussians. Inaccuracies in the monocular depth estimates further amplify the problem. The original density control strategy is ineffective at removing such Gaussians. Instead, through densification, it often transforms them into small isolated islands of Gaussians that impact the final reconstruction quality.

We found the clustering-based pruning strategy to be effective at removing such islands of Gaussians at an early stage during the optimization. We specifically employ the 90th percentile of the $k$-distance distribution to determine the DBSCAN radius $\epsilon$. A 90th percentile threshold ensures that $\epsilon$ remains sufficiently tight to prevent these islands from merging with the main cluster. Although this may result in over-pruning some of the true geometry near the borders in case the 90th percentile distance is small, these regions are typically recovered during subsequent iterations through densification. We observed that choosing an $\epsilon$ lower than the 90th percentile distance could sometimes result in the pruning of flat, textureless surfaces, since the optimization naturally favors a sparse distribution of larger Gaussians to represent such regions.

\begin{figure}[!htb]
    \centering
    \includegraphics[width=0.48\linewidth]{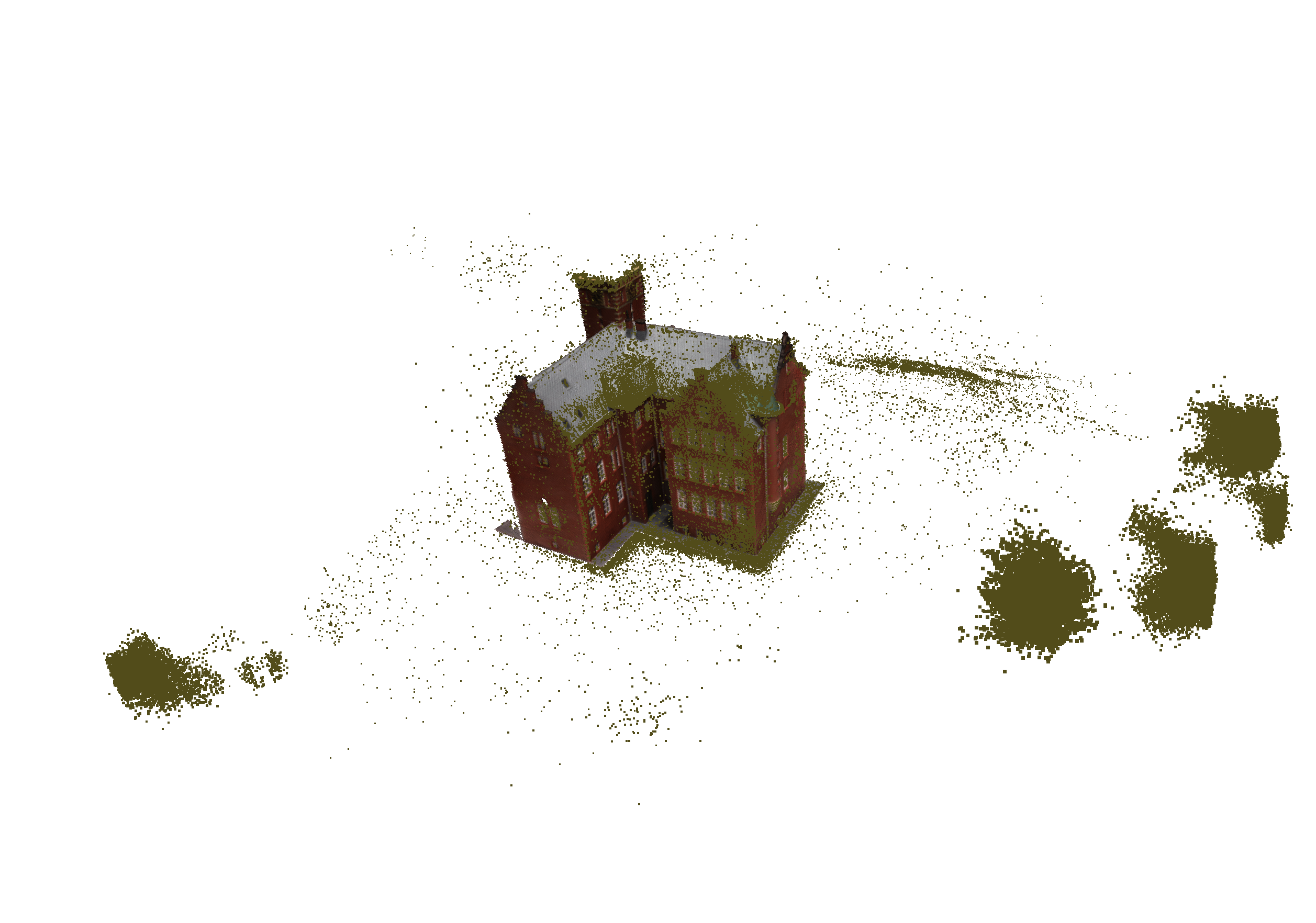}\hfill
    \includegraphics[width=0.48\linewidth]{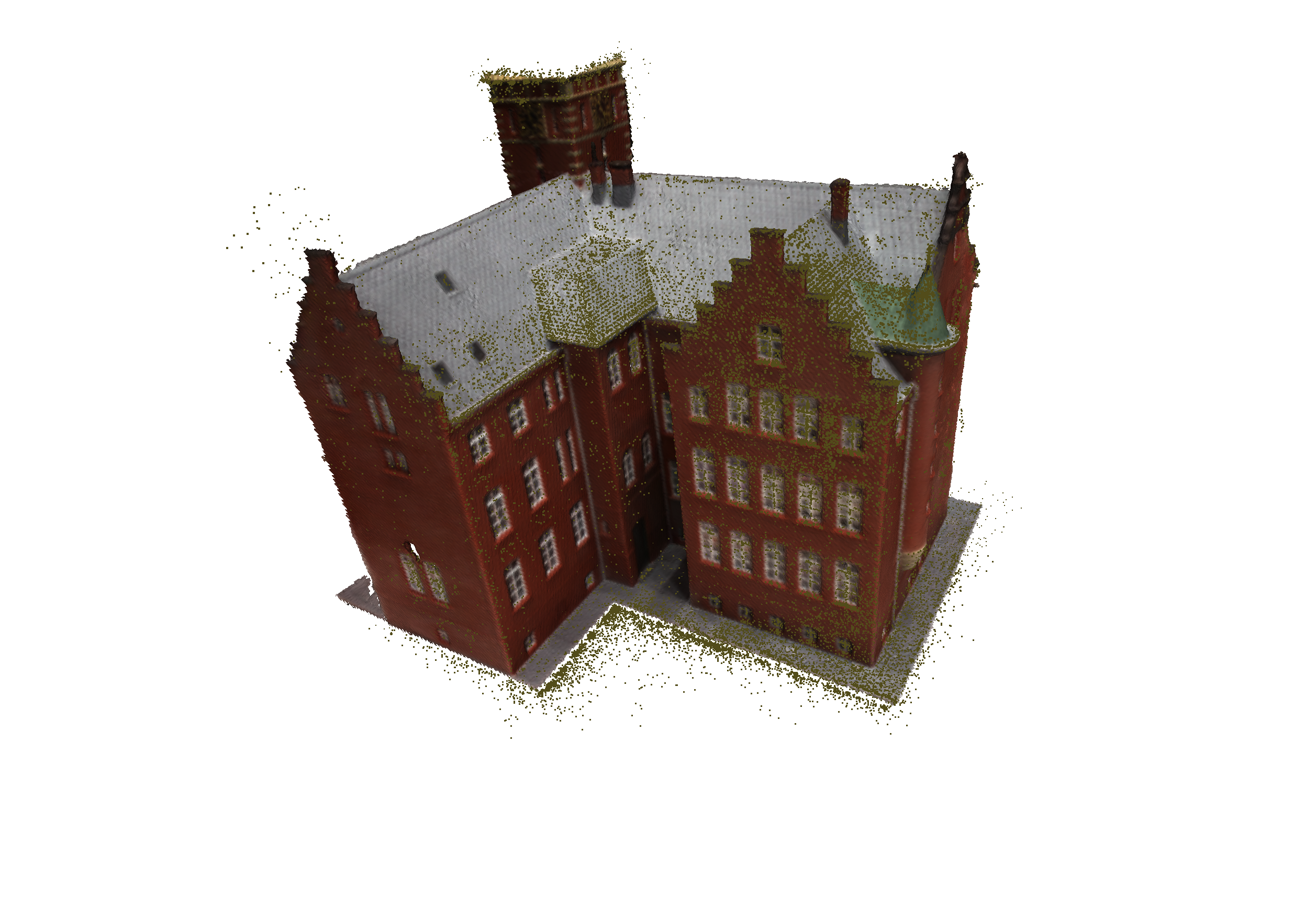}
    \caption{\textbf{Visualization of Gaussians before (left) and after (right) cluster-based pruning.} Each dot represents the mean of the corresponding 2D Gaussian.}
    \label{fig:cluster_pruning}
\end{figure}

Figure~\ref{fig:cluster_pruning} illustrates the effectiveness of the proposed method. Gaussians close to the surface of the object survive, while the others are eliminated. Our pruning strategy is also robust to any geometry hallucinated by the monocular depth estimator as long as the object in focus remains the largest cluster. Mesh reconstruction is also possible in the absence of object masks, provided that sufficient multi-view images of the object are supplied. However, there is a slight reduction in fine-grained quality compared to masked training.

\section{Additional Visualizations}

\begin{figure}[!htb]
    \centering
    \begin{subfigure}[b]{0.32\linewidth}
        \centering
        \includegraphics[width=\linewidth]{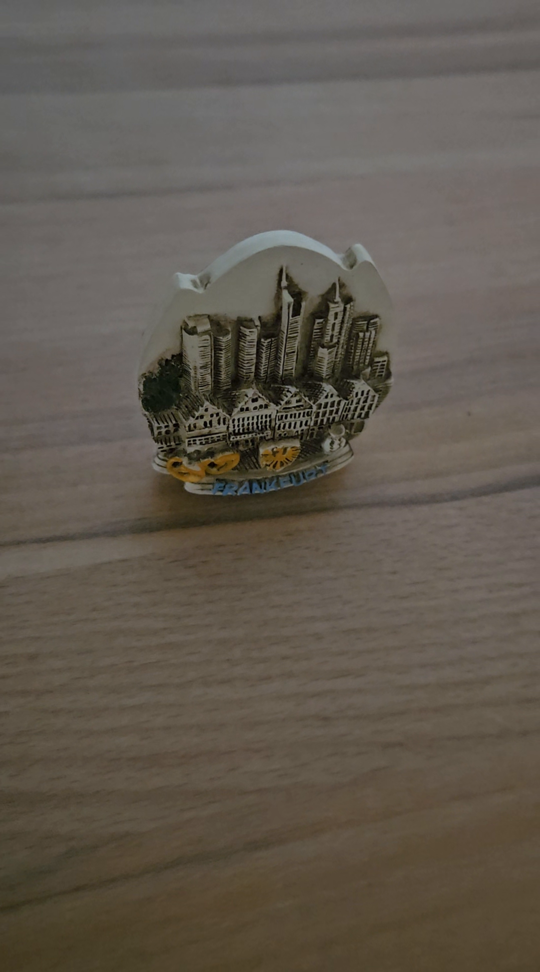}
        \caption{Ground Truth}
    \end{subfigure}
    \hfill
    \begin{subfigure}[b]{0.32\linewidth}
        \centering
        \includegraphics[width=\linewidth]{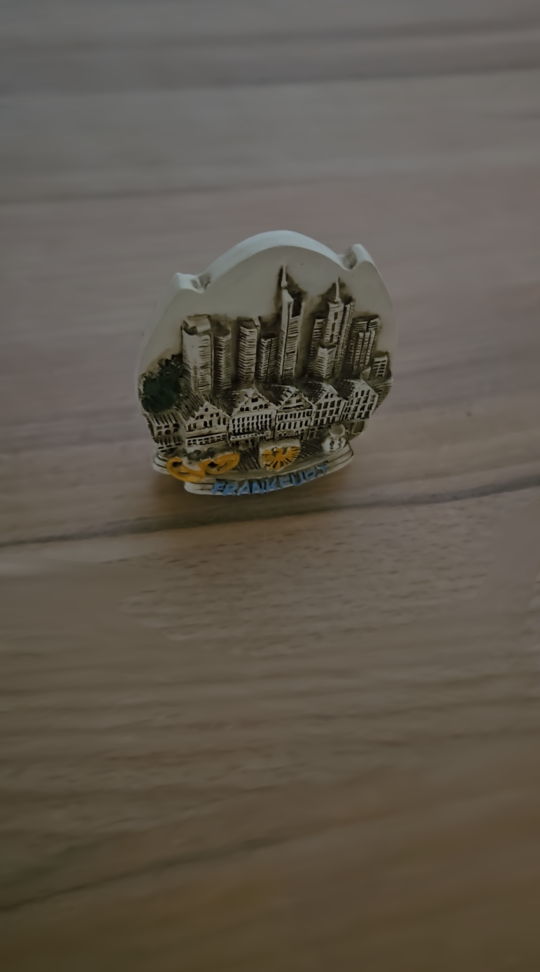}
        \caption{2D-SuGaR}
    \end{subfigure}
    \hfill
    \begin{subfigure}[b]{0.32\linewidth}
        \centering
        \includegraphics[width=\linewidth]{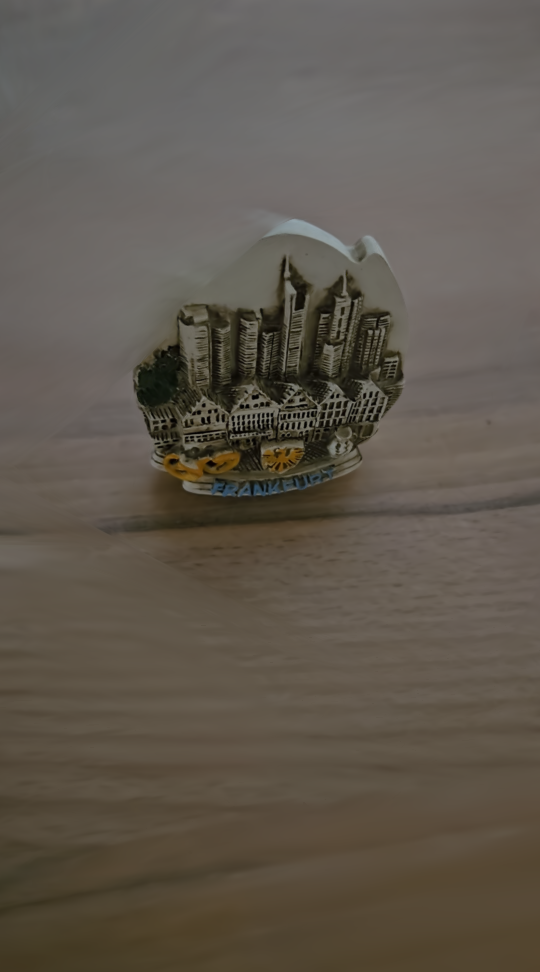}
        \caption{2DGS}
    \end{subfigure}

    \caption{\textbf{Novel View Synthesis on a custom scene with challenging SfM initialization.} Left: A ground truth RGB image from the scene. Middle: NVS results from our method. Right: NVS results from 2DGS.}
    \label{fig:frankfurt_nvs}
\end{figure}

In order to emphasize the flexibility provided by depth-based Gaussian initializations, we construct a custom scene for which it is challenging to get a good SfM initialization. The scene consists of several images of a small fridge magnet placed on a large, flat and smooth table. Figure~\ref{fig:frankfurt_nvs} illustrates one such image from the scene, along with the corresponding novel view synthesis. While our method results in a realistic rendering of the scene, 2DGS struggles to recover from the poor SfM initialization and fails to recreate the scene without artifacts.

\begin{figure*}[!htbp]
  \centering

  {\textbf{2D-SuGaR}}
  \vspace{0.5em}
  \includegraphics[
    width=\textwidth,
    height=0.43\textheight,
    keepaspectratio
  ]{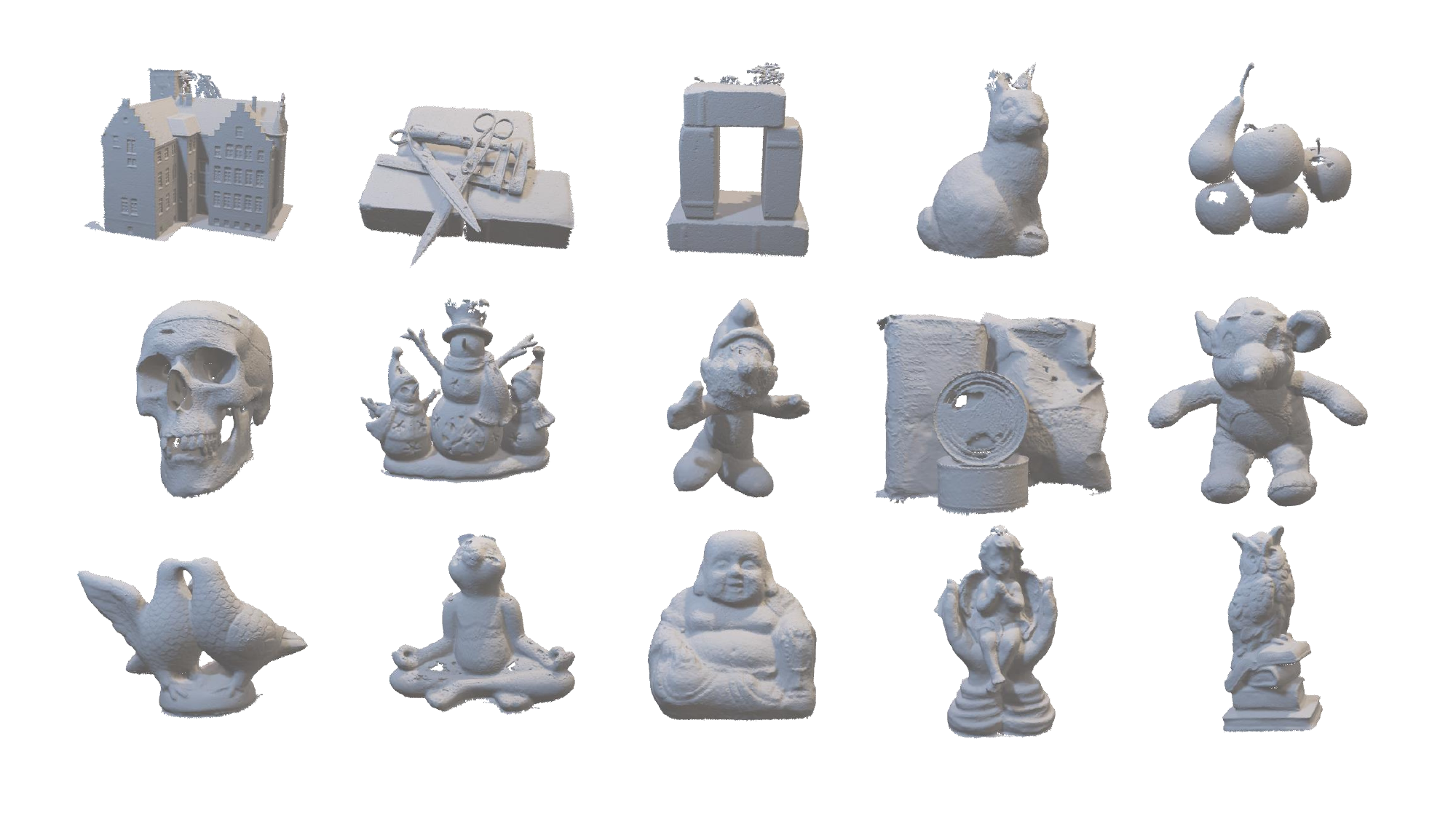}

  \vspace{0.8em}
  \hrule
  \vspace{0.8em}

  {\textbf{2DGS}}
  \vspace{0.5em}
  \includegraphics[
    width=\textwidth,
    height=0.43\textheight,
    keepaspectratio
  ]{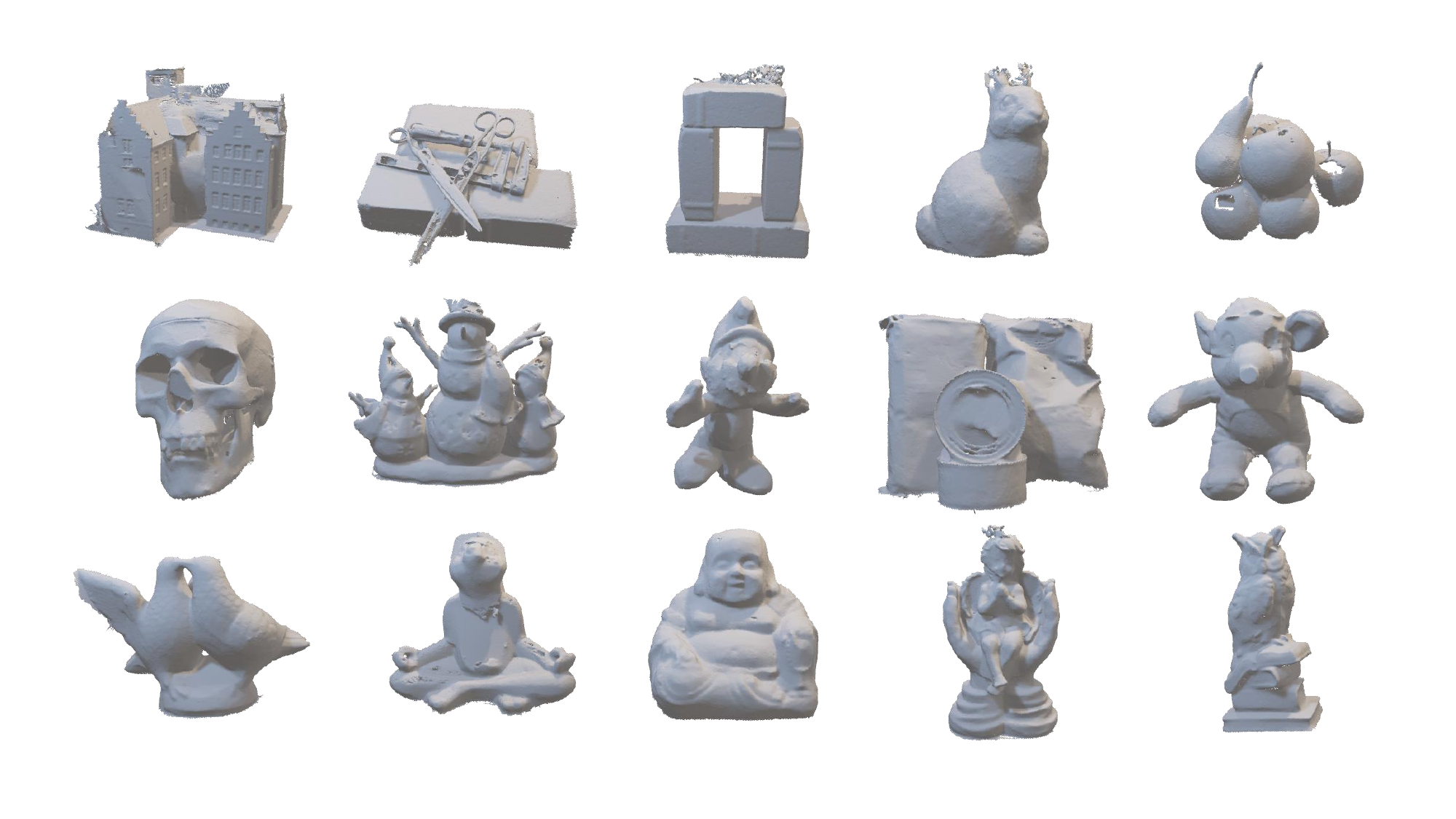}

  \caption{
    \textbf{Qualitative comparison of DTU mesh reconstructions between our method (top) and 2DGS (bottom).}
  }
  \label{fig:mesh_2dsugar_vs_2dgs}
\end{figure*}

Figure~\ref{fig:mesh_2dsugar_vs_2dgs} visualizes several meshes from the DTU dataset obtained using our method and 2DGS. When compared to 2DGS, our method better preserves curvature of surfaces, fine textures and results in smoother reconstructions of the underlying surface.

\section{Appearance Reconstruction}
While our method prioritizes geometric accuracy, we also evaluate rendering quality to ensure photorealistic appearance is preserved. The refined meshes are rendered using the 3D Gaussian rasterizer. As shown in Table~\ref{tab:refinement_psnr}, our method without refinement achieves a PSNR of 34.23 dB, which is competitive with 2DGS (34.38 dB) and SuGaR (34.57 dB), while significantly outperforming all methods in geometric accuracy (CD: 0.72). Although 3DGS achieves the highest PSNR (35.76 dB), its geometric reconstruction is substantially worse (CD: 1.96). Notably, our approach achieves the fastest training time (10.5 minutes), making it both accurate and efficient. This demonstrates that our robust initialization and training strategy successfully balances geometric fidelity with appearance quality, without sacrificing rendering performance for improved surface reconstruction. Note that it is also possible to optimize UV textures on the mesh using differentiable mesh rendering. While this approach may result in marginal reduction in the rendering metrics compared to the Gaussians-based rendering, it offers full compatibility with classical computer graphics workflows and graphics engines.

\begin{table}[!htb]
\small
\centering
\begin{tabular}{@{}lccc@{}}
\toprule
Method  & CD$\downarrow$ & PSNR $\uparrow$ & Time $\downarrow$ \\
\midrule
Ours w/o refinement & \textbf{0.72} &  {34.23} & \textbf{10.5m}\\
\midrule
3DGS  & 1.96 & \textbf{35.76} & 11.2m\\
Sugar & 1.33 & {34.57} &  65m \\
2DGS  & 0.76 & 34.38 & 10.9m \\

\bottomrule
\end{tabular}
\caption{\textbf{Performance comparison on DTU dataset.} We report the averaged chamfer distance, PSNR (training-set view) and reconstruction time}
\label{tab:refinement_psnr}
\end{table}
 
\bibliographystyle{EG-style/eg-alpha-doi} 
\bibliography{main}       

